\pdfoutput=1

\documentclass[11pt]{article}

\usepackage[final]{acl}

\usepackage{times}
\usepackage{latexsym}
\usepackage{url} 
\usepackage[T1]{fontenc}
\usepackage{tcolorbox}
\usepackage{graphicx}
\usepackage{xspace}

\newcommand{\modelName}{\textsc{UnityAI-Guard}\xspace}

\usepackage[utf8]{inputenc}

\usepackage{microtype}

\usepackage{inconsolata}

\newcommand{\demo}{\raisebox{-1.5pt}{\includegraphics[height=1.05em]{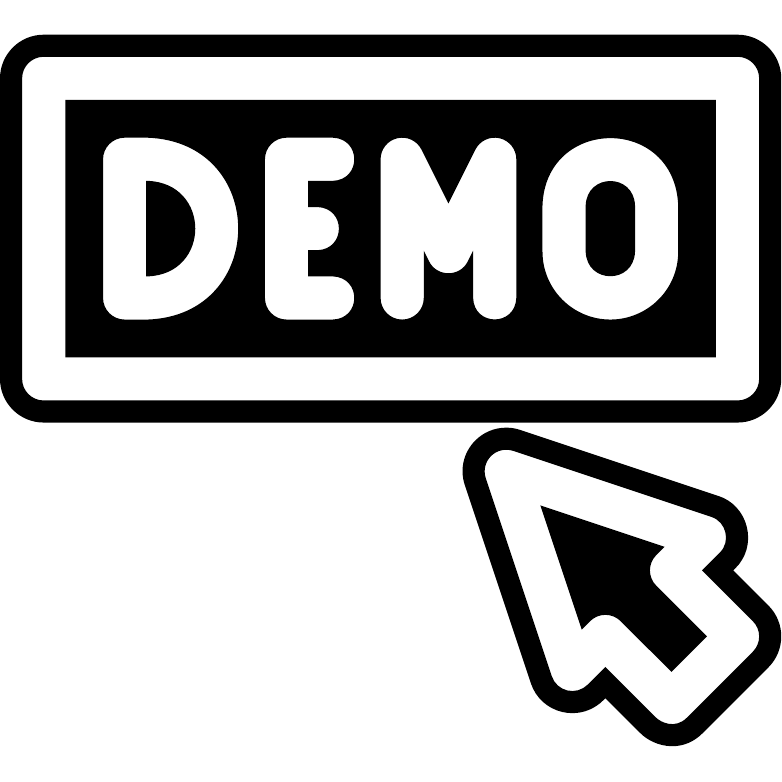}}\xspace}

\newcommand{\web}{\raisebox{-1.5pt}{\includegraphics[height=1.05em]{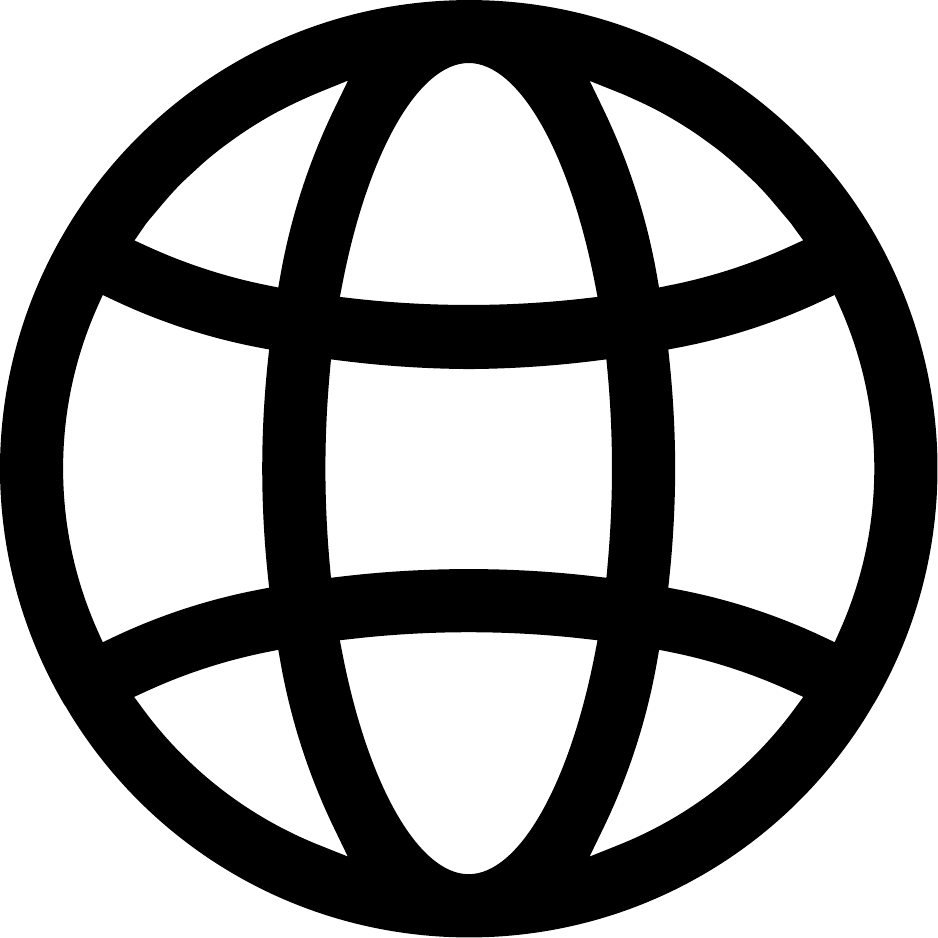}}\xspace}
\NewDocumentCommand\emojismile{}{
    \includegraphics[scale=0.05]{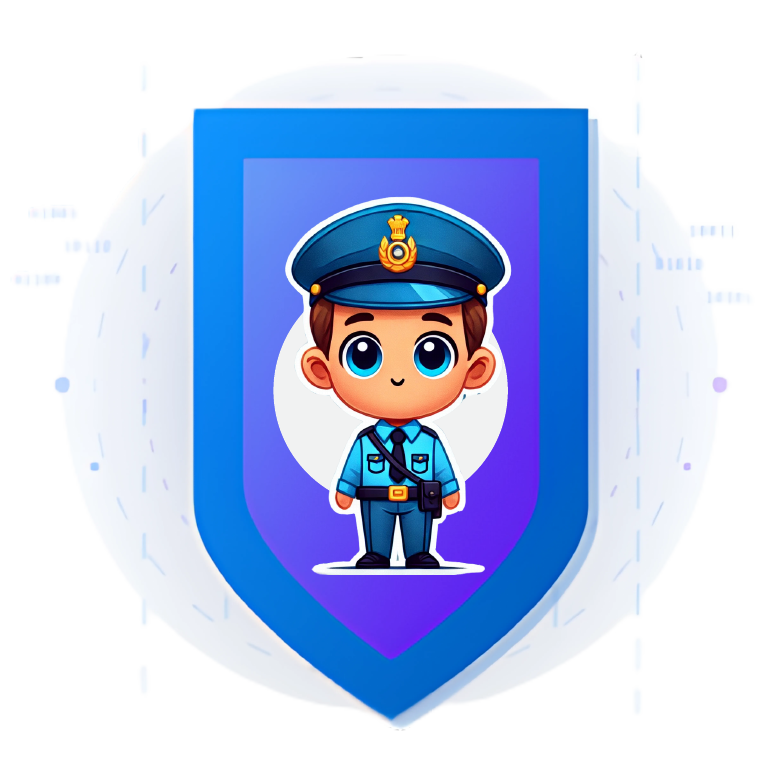}
}
\usepackage{graphicx}

\title{\modelName \emojismile: Pioneering Toxicity Detection Across Low-Resource Indian Languages \\ 
\small \textcolor{red}{WARNING: The content contains samples that are offensive and toxic.}}

\author{
 \textbf{Himanshu Beniwal\textsuperscript{$\heartsuit$*}},
 \textbf{Reddybathuni Venkat\textsuperscript{$\heartsuit$*}},
 \textbf{Rohit Kumar\textsuperscript{$\clubsuit$*}},
\\
\textbf{Birudugadda Srivibhav\textsuperscript{$\heartsuit$*}},
 \textbf{Daksh Jain\textsuperscript{$\heartsuit$*}},
 \textbf{Pavan Doddi\textsuperscript{$\heartsuit$}},
 \textbf{Eshwar Dhande\textsuperscript{$\heartsuit$}},
\\
 \textbf{Adithya Ananth\textsuperscript{$\diamond$}},
\textbf{Kuldeep\textsuperscript{$\heartsuit$}},
 \textbf{Mayank Singh\textsuperscript{$\heartsuit$}}
\\
\\
 \textsuperscript{$\heartsuit$}Indian Institute of Technology Gandhinagar,
 \textsuperscript{$\clubsuit$}Indian Institute of Technology Goa, \\
 \textsuperscript{$\diamond$}Indian Institute of Technology Tirupati
 \\
 \small{
   \textbf{Correspondence:} \href{mailto:lingo@iitgn.ac.in}{lingo@iitgn.ac.in}
 }
}

\begin{document}
\maketitle
\begin{abstract}
This work introduces \modelName, a framework for binary toxicity classification targeting low-resource Indian languages. While existing systems predominantly cater to high-resource languages, \modelName addresses this critical gap by developing state-of-the-art models for identifying toxic content across diverse Brahmic/Indic scripts. Our approach achieves an impressive average F1-score of 84.23\% across six languages, leveraging a dataset of 567k training instances and 30k manually verified test instances. By advancing multilingual content moderation for linguistically diverse regions, \modelName also provides public API access to foster broader adoption and application.
\begin{center}
\begin{tabular}{rc}
    \web-Website & \href{https://lingo.iitgn.ac.in/unityai-guard/}{UnityAI-Guard} \\
    \demo-Demo & \href{https://bit.ly/UnityAI-Guard}{bit.ly/UnityAI-Guard}\\
\end{tabular}
\end{center}
\def\thefootnote{*}\footnotetext{Equal Contribution}\def\thefootnote{\arabic{footnote}}

\end{abstract}

\section{Introduction}
India represents one of the most linguistically diverse regions in the world, with 22 officially recognized languages and hundreds of dialects \citep{singh2020asroil, indictrans2}. Among these, Hindi leads with approximately 528 million speakers, while Telugu (81 million), Marathi (83 million), Punjabi (33 million), and Urdu (51 million) each represent significant linguistic communities\footnote{\url{https://en.wikipedia.org/wiki/List_of_languages_by_number_of_native_speakers_in_India}} \citep{indicnlpsuite, lowresource}. 
Despite their large speaker bases, most of these languages remain computationally low-resourced, especially in specialized NLP applications like content moderation \citep{dongare2024creating, narayan2023hate, hegde-etal-2023-mucs}. This has created vibrant low-resource language users but simultaneously presents significant challenges in content moderation and user safety \citep{toxicitycosta, ptp, rahman-etal-2024-binary}.

\begin{figure}[t]
    \centering
    \includegraphics[width=\linewidth]{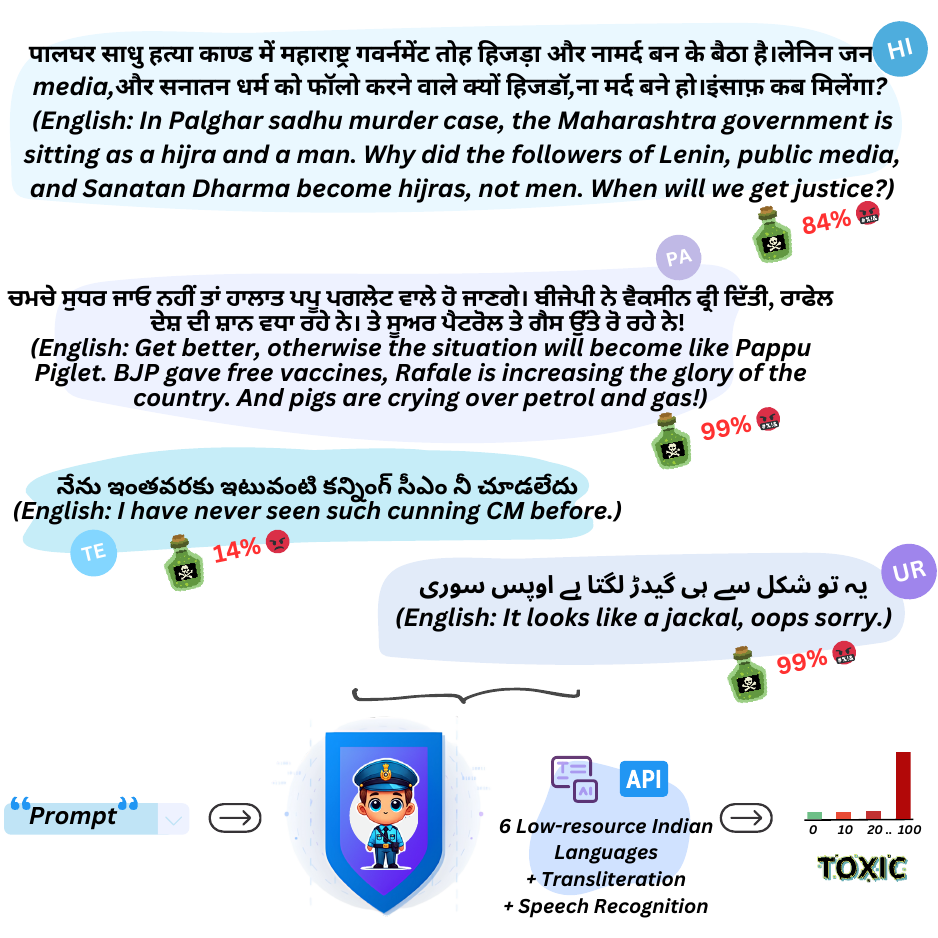}
    \caption{Different examples of toxic prompts in four different scripts. \textbf{\textit{Takeaway}}: \textit{\modelName detects toxicity in six languages, transliteration for ease, and speech recognition for better usability}.}
    \label{fig:examples}
\end{figure}

\begin{table}[t]
\centering
\begin{tabular}{ccccc}
\textbf{} & \multicolumn{2}{c}{\textbf{Train}} & \multicolumn{2}{c}{\textbf{Test}} \\ \hline
\textbf{Lang.} & \textbf{Tox.} & \textbf{Neu.} & \textbf{Tox.} & \textbf{Neu.} \\ \hline
\textit{Hindi} & 39691 & 32473 & 2500 & 2500 \\
\textit{Telugu} & 62309 & 95282 & 2500 & 2500 \\
\textit{Marathi} & 18092 & 20079 & 2500 & 2500 \\
\textit{Urdu} & 32690 & 32690 & 2500 & 2500 \\
\textit{Punjabi} & 18186 & 16159 & 2648 & 2352 \\
\textit{Tamil} & 100000 & 100000 & 2500 & 2500 \\ \hline
\end{tabular}%

\caption{The dataset split and the balance for the toxic and neutral pairs. Note that ``Tox'' and ``Neu'' represent the number of toxic and neutral instances, respectively. \textit{\textbf{Takeaway}}: \textit{We keep at least 5k manually-verified test pairs for the evaluation}. }
\label{tab:dataset-split}
\end{table}

\begin{figure}[t]
    \centering
    \includegraphics[width=0.7\linewidth]{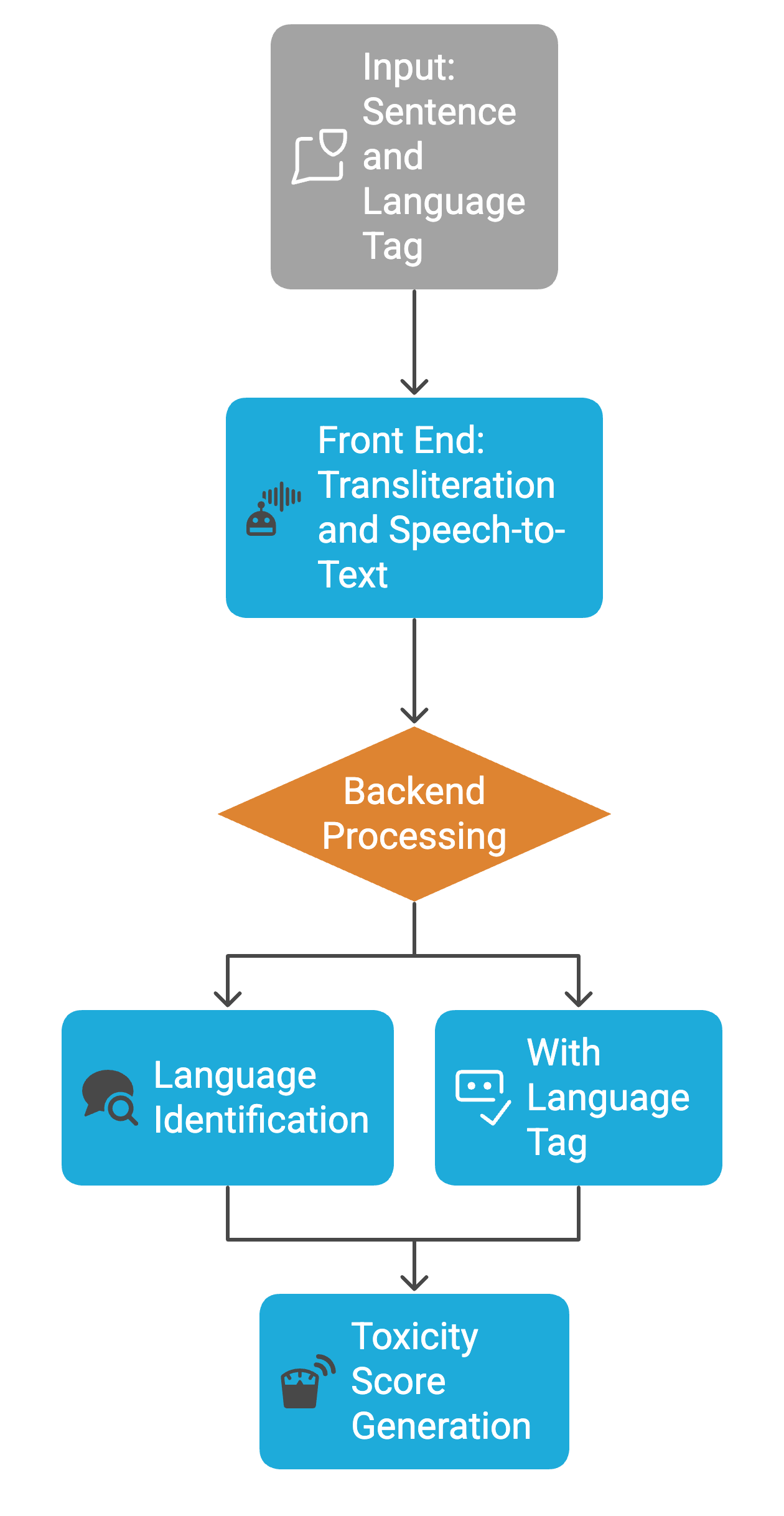}
    \caption{Architectural view for \modelName. \textbf{\textit{Takeaway}}: \textit{Front-end has speech-to-text and transliteration, whereas backend detects toxicity with/without language tags}.}
    \label{fig:archi}
\end{figure}

Online toxicity—including hate speech, abusive language, and harassment—has emerged as a critical concern across social platforms \citep{aodhora-etal-2025-cuet}. These languages face multiple constraints that hinder effective NLP implementations: limited digitized corpora, scarce annotated datasets for toxicity, inadequate linguistic resources, and minimal pre-trained language models \citep{maity-etal-2024-toxvidlm, indicllmsuite, indictrans2, lekshmiammal-etal-2022-nitk}. Furthermore, linguistic phenomena such as code-mixing with English and other regional languages, script variations (particularly in Urdu), and dialectal differences create additional complications for toxicity detection systems \citep{saeed2021roman}. Lastly, a popular tool such as ``\textit{Perspective-API}''\footnote{\url{https://perspectiveapi.com/}} is still available in 17 widely spoken languages and only supports Hindi and Hinglish, not other Indian languages. However, even \texttt{llama-guard-3-8B} \citep{llama3} only supports Hindi as the Indian language, out of its support for eight languages. 

To address this critical research gap, we introduce \modelName, a comprehensive framework that supports six languages widely spoken in the Indian subcontinent with multiple functionalities: toxicity classification, transliteration capabilities, speech recognition integration, and programmatic access through API endpoints. We define the entire architecture in Figure~\ref{fig:archi}, where we illustrate the front-end and back-end of the framework. More details in Section~\ref{sec:archi}.

\paragraph{Contributions} Our contributions can be summarized as follows:
\begin{itemize}
    \item We introduce state-of-the-art binary toxicity classification models across six languages with diverse scripts.
    \item We created the largest toxicity dataset comprising 567,651 training instances and 30k test instances that underwent rigorous quality assurance through manual verification of all samples by two native speaker annotators per language, ensuring authentic content rather than synthetically generated data.
    \item We present a comprehensive web portal equipped with accessibility features, including transliteration support, speech recognition capabilities, and programmatic access via API endpoints.
\end{itemize}

\section{Related Works}

\subsection{Content Moderation Datasets}
Content moderation has been extensively studied in high-resource languages such as English, German, and French, with seminal work by \citet{ye-etal-2023-multilingual} and \citet{wang-etal-2025-stand} establishing foundational datasets for detecting hate speech in English. For non-English European languages, \citet{farhan-2025-hyderabadi} presented cross-lingual approaches for Italian and Spanish moderation.

In contrast to high-resource languages, content moderation for Indian languages has received comparatively limited attention. \citet{bohra-etal-2018-dataset} and \citet{srivastava-2025-dweshvaani} explored the detection of hate speech in Hindi-English code-mixed text, while \citet{garain-etal-2021-junlp}, \citet{raihan-etal-2023-offensive}, and \citet{kedia-nandy-2021-indicnlp} presented benchmark datasets for the identification of offensive languages in Dravidian languages.

\subsection{Existing Systems and Tools for Content Moderation in Indian Languages}
A popular tool named ``\textit{Perspective-API}'' has gained significant traction but restricts support to only Hindi and Hinglish. \citet{indicnlpsuite} introduced the IndicNLPSuite and \citet{indicllmsuite} presented Indic-align; however, both approaches rely on synthetically generated toxic content, which fails to capture the complex nuances of naturally occurring harmful language. The previous works by \citet{saikh-etal-2024-emojis} and \citet{raihan-etal-2023-offensive} focus on fine-tuned models; they lack support for other languages and do not provide the UI and API-support \citep{saeed2021roman, lekshmiammal-etal-2022-nitk, kedia-nandy-2021-indicnlp}.

\begin{figure}[t]
    \centering
    \includegraphics[width=\linewidth]{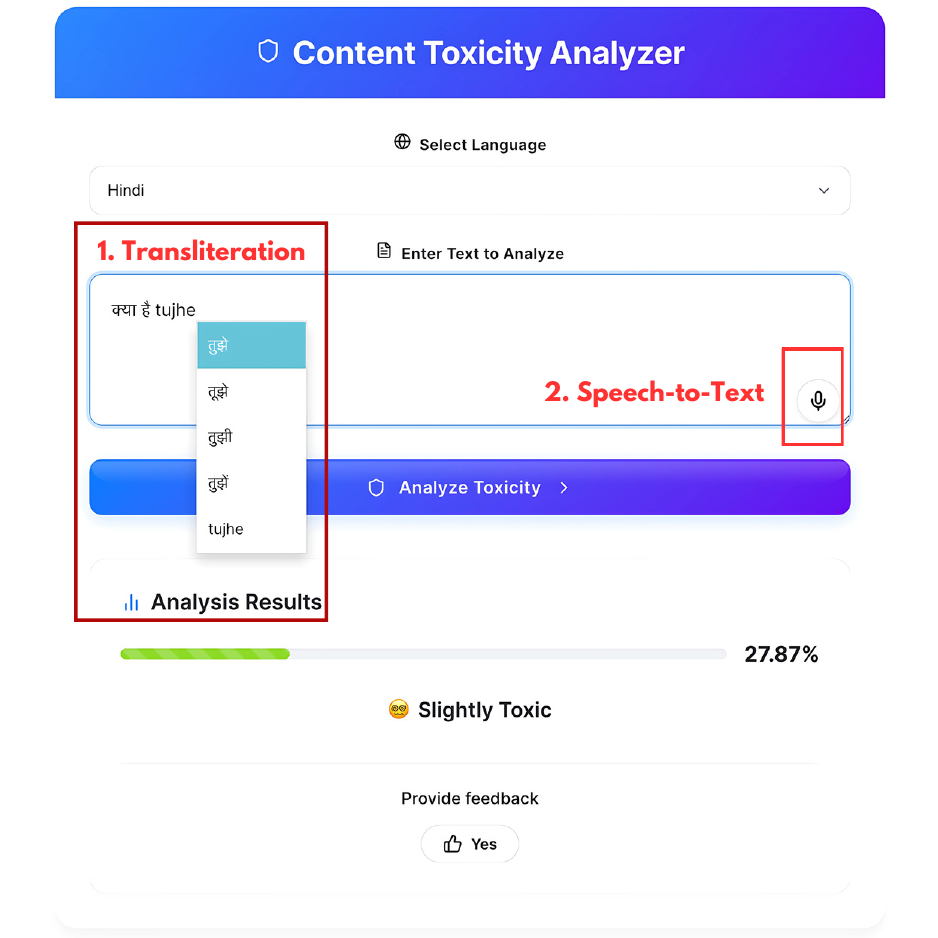}
    \caption{Two major add-ons for the UI improvment. \textbf{\textit{Takeaway}}: \textit{We present transliteration and speech-to-text for the non-native speakers}.}
    \label{fig:trans-speech}
\end{figure}

\subsection{Research Limitations and Our Contributions}
Previous research efforts in content moderation for Indian languages have been constrained by limited language coverage and insufficient manually verified data \citep{indictrans2, dongare2024creating, kedia-nandy-2021-indicnlp, rahman-etal-2024-binary}.  To address this limitation, we propose comprehensive datasets and models for low-resource Indian languages with particular emphasis on diverse scripts, including Tamil, Punjabi, and Marathi, which have remained largely unexplored in previous research efforts. Our work addresses these limitations by contributing a substantial corpus of manually inspected samples across multiple low-resource languages. We also conduct a comparative analysis of classification performance across models of varying capacity.

Our contribution represents the first large-scale dataset with manual verification of test samples, establishing a more reliable benchmark for evaluating content moderation models across Indian languages. Additionally, our work provides users and developers with an easy-to-use setup and comprehensive API support.

\section{Experiments}
\subsection{Dataset}

For dataset construction, we curated online sources\footnote{The dataset is curated from News Channels, Keyword-based-searches, and licensed sources only. More details in the Appendix~\S\ref{sec:appendix-dataset}} for six low-resource Indian languages: Hindi (hi), Telugu (te), Marathi (mr), Urdu (ur), Tamil (ta), and Punjabi (pa).
The selection criteria were based on resource scarcity and script diversity, as each language employs a distinctly different writing system. In addition, we account for the transliteration and code-mixing phenomena between these languages and English. The distribution of toxic and neutral pairs between the train test splits is presented in Table~\ref{tab:dataset-split}. Note that given the budget constraints, only the test set (30,000 instances) was formulated by manual inspection, where we had 567,651 instances in the training set. We plan to release the dataset under the CC-BY-4.0 license.

\paragraph{Quality Control} The dataset underwent annotation by two native speakers of each language. These annotators were unpaid volunteer students who participated in curating a comprehensive toxicity-labeled dataset over a 30-day period. Annotators were instructed to classify content as toxic if it aligned with any of the categories defined by \texttt{Llama-Guard-3-8B}\footnote{\url{https://huggingface.co/meta-llama/Llama-Guard-3-8B}}. Additionally, we implemented a keyword-based approach whereby content containing specific toxic terms was systematically categorized as toxic.

\subsection{Models}
To demonstrate variation in understanding low-resource languages with diverse scripts, we employ three differently sized models: \textit{\textbf{(1)}} \texttt{mbert-base-uncased} (560M parameters, \citep{bert}), \textbf{\textit{(2)}} \texttt{llama-3.2-1B} (1B parameters, \citep{llama3}), and \textbf{\textit{(3)}} \texttt{aya-expanse-8B} (8B parameters, \citep{aya8b}). This selection represents a strategic gradient of model capacities, allowing us to analyze how parameter count correlates with cross-lingual understanding in Indic languages. We train the models with the classification task, where text is passed as the input and label as the output.

We present the zero-shot evaluation results in Table~\ref{tab:zs-scores}. The models demonstrate significantly diminished performance in zero-shot settings. For these experiments, inference was conducted by simply providing the input sentence and using the generated output as the predicted label. We aim to release the models under the MIT license.

\subsection{Metrics}
For evaluation, we utilize standard classification metrics, including Precision (P), Recall (R), F1 score (F1), and Accuracy (A), to assess model performance across the six target languages.

\begin{table*}[t]
\resizebox{\textwidth}{!}{
\begin{tabular}{ccccccccccccc}
\textbf{} & \multicolumn{4}{c}{\textbf{\texttt{mBERT}}} & \multicolumn{4}{c}{\textbf{\texttt{llama-3.2-1B}}} & \multicolumn{4}{c}{\textbf{\texttt{aya-expanse-8b}}} \\ \hline
\textbf{Language} & \textbf{P} & \textbf{R} & \textbf{F1} & \textbf{A} & \textbf{P} & \textbf{R} & \textbf{F1} & \textbf{A} & \textbf{P} & \textbf{R} & \textbf{F1} & \textbf{A} \\ \hline
\textit{Hindi} & 83.04\ & 81.97\ & 82.50\ & 82.63\ & 83.81\ & 84.21\ & 84.01\ & 83.99\ & 90.30\ & 82.05\ & 85.98\ & 86.63\ \\
\textit{Telugu} & 93.23\ & 83.12\ & 87.88\ & 88.54\ & 93.69\ & 82.52\ & 87.75\ & 88.48\ & 96.72\ & 79.04\ & 86.99\ & 88.18\ \\
\textit{Marathi} & 85.47\ & 85.23\ & 85.35\ & 85.38\ & 89.81\ & 88.80\ & 89.31\ & 89.37\ & 91.05\ & 87.36\ & 89.17\ & 89.39\ \\
\textit{Urdu} & 78.92\ & 85.04\ & 81.87\ & 81.17\ & 86.40\ & 81.75\ & 84.01\ & 84.45\ & 84.32\ & 91.03\ & 87.55\ & 87.06\ \\
\textit{Punjabi} & 77.15\ & 81.71\ & 79.37\ & 77.43\ & 81.40\ & 81.52\ & 81.46\ & 80.29\ & 81.73\ & 80.65\ & 81.19\ & 80.14\ \\
\textit{Tamil} & 80.08\ & 83.48\ & 81.75\ & 81.36\ & 76.15\ & 81.88\ & 78.91\ & 78.12\ & 83.79\ & 74\ & 78.59\ & 79.84\
\\ \hline
\end{tabular}
}
\caption{Accuracy scores (in percentage) over different languages using the three \textit{fine-tuned} models. We report performance metrics as Precision (P), Recall (R), Accuracy (A), and F1. \textbf{\textit{Takeaway}}: \textit{We observe that the smaller and larger LMs are very comparable}.}
\label{tab:scores}
\end{table*}

\begin{table*}[t]
\resizebox{\textwidth}{!}{
\begin{tabular}{crrrrccccllll}
\textbf{} & \multicolumn{4}{c}{\textbf{\texttt{mBERT}}} & \multicolumn{4}{c}{\textbf{\texttt{llama-3.2-1B}}} & \multicolumn{4}{c}{\textbf{\texttt{aya-expanse-8b}}}\\ \hline
\textbf{Language} & \multicolumn{1}{c}{\textbf{P}} & \multicolumn{1}{c}{\textbf{R}} & \multicolumn{1}{c}{\textbf{F1}} & \multicolumn{1}{c}{\textbf{A}} & \textbf{P} & \textbf{R} & \textbf{F1} & \textbf{A} & \multicolumn{1}{c}{\textbf{P}} & \multicolumn{1}{c}{\textbf{R}} & \multicolumn{1}{c}{\textbf{F1}} & \multicolumn{1}{c}{\textbf{A}} \\ \hline
\textit{Hindi} & 50.85 & 86.89 & 64.16 & 52.14 & 48.52 & 23.75 & 31.89 & 49.31 & 69.09 & 78.81 & 73.63 & 71.80 \\
\textit{Telugu} & 52.85 & 97.40 & 68.52 & 55.26 & 54.03 & 22.76 & 32.03 & 51.70 & 59.13 & 52.08 & 55.38 & 58.04 \\
\textit{Marathi} & 50.72 & 95.91 & 66.35 & 51.40 & 53.24 & 31.90 & 39.90 & 51.98 & 59.98 & 70.79 & 64.94 & 61.81 \\
\textit{Urdu} & 51.40 & 97.10 & 67.22 & 52.67 & 53.80 & 26.73 & 35.72 & 51.90 & 63.69 & 60.15 & 61.87 & 62.94 \\
\textit{Punjabi} & 53.67 & 87.85 & 66.63 & 53.25 & 54.19 & 55.28 & 54.73 & 51.41 & 68.05 & 32.34 & 43.84 & 55.98 \\
\textit{Tamil} & 57.00 & 41.68 & 48.15 & 55.12 & 5.06 & 1 & 1.67 & 41.12 & 29.71 & 14.20 & 19.22 & 40.30 \\ \hline
\end{tabular}
}
\caption{Accuracy scores (in percentage) over different languages using the three \textit{Zero-Shot} models. We report performance metrics as Precision (P), Recall (R), Accuracy (A), and F1. \textbf{\textit{Takeaway}}: \textit{The zero-shot models are performing really poor}.}
\label{tab:zs-scores}
\end{table*}

The experiments demonstrate that \texttt{aya-expanse-8b} achieves superior accuracy (87.21\%) compared to \texttt{mbert-base} (83.40\%), as detailed in Table~\ref{tab:scores}. Similarly, we observe a consistent improvement in F1 scores as the model size increases, with values of 83.67\%, 84.28\%, and 86.96\%, respectively. These results indicate that larger models demonstrate enhanced capability in processing and understanding diverse Indian scripts than the zero-shot models. 

\section{System Architecture}
\label{sec:archi}
\subsection{Front-End}
We have designed an intuitive user interface that enables experimentation with our language-specific models. As illustrated in Figure~\ref{fig:trans-speech}, the system incorporates supplementary functionalities: \textit{\textbf{(1)}} \textit{Transliteration}\footnote{\url{https://www.npmjs.com/package/react-transliterate}} capabilities to facilitate script conversion, and \textbf{\textit{(2)}} \textit{Speech-to-Text}\footnote{\url{https://developer.mozilla.org/en-US/docs/Web/API/Web_Speech_API}} functionality to enhance user accessibility.

\begin{figure}[t]
    \centering
    \includegraphics[width=\linewidth]{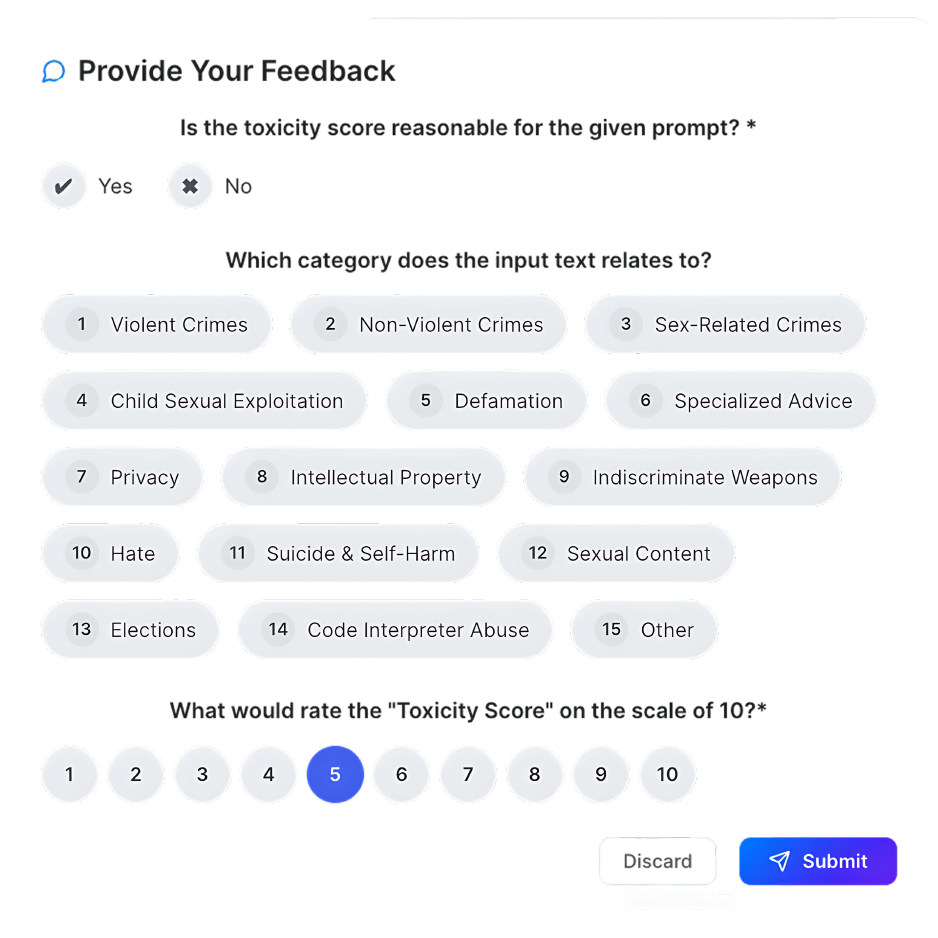}
    \caption{Feedback form over the user's input and model's output. \textbf{\textit{Takeaway}}: \textit{We capture the feedback over the input and output using the Google Sheets integration}.}
    \label{fig:feedback}
\end{figure}

\paragraph{Feedback Collection}
We also collect optional user feedback in a structured form as illustrated in Figure~\ref{fig:feedback}. The feedback mechanism captures multiple evaluation dimensions, including the user's input text, language identification, assigned toxicity score, score relevance assessment, user-selected content category, and a quantitative rating of model response quality on a 10-point scale. This data is systematically archived via Google Sheets integration for subsequent analysis. To maintain methodological consistency with established benchmarks as implemented in \texttt{llama-guard-3-8B} \citep{llama3}, we adopt the same 14 taxonomic categories for content classification.

\subsection{Back-End}

We also deploy the trained models using FastAPI\footnote{\url{https://github.com/fastapi/fastapi/}}. 
We show a sample code snippet to call the models by passing the input prompt and the language tag. It also showcases the output in a JSON format with the toxicity score and confidence score\footnote{More details are available here: \url{https://lingo.iitgn.ac.in/unityai-guard/\#/docs}}. We keep the rate limit of 60 requests per minute; after that, the message with the rate limit is popped on. 

\begin{tcolorbox}[title=Code Snippet for API]
\begin{verbatim}
import requests

headers = {
    "X-API-Key": "your_api_key_here",
    "Content-Type": "application/json"
}

data = {
    "text": "tu bohot ganda hai",
    "lang": "hi"
}

response = requests.post(
    "https://lingo.iitgn.ac.in/
    unityai-guard/api", 
    headers=headers,
    json=data
)

print(response.json())

# Output: {
  "confidence": 78.24, "is_toxic": 
  true, "toxicity": 21.76
}
\end{verbatim}
\end{tcolorbox}

\section{Conclusion and Future Directions}
We presented \modelName, a comprehensive framework for toxicity detection across six low-resource Indian languages with diverse scripts. Our contributions include the largest annotated toxicity dataset for Hindi, Telugu, Marathi, Urdu, Punjabi, and Tamil, validated by two native speakers, and an accessible web portal with transliteration, speech recognition, and API capabilities. Experimental results demonstrate the efficacy of our approach across different model capacities. Future work will focus on expanding language coverage to include additional Indian languages, mainly from Dravidian and North-Eastern families, incorporating contextual understanding for culturally-specific nuances and developing robust cross-lingual transfer approaches to accommodate India's script diversity. We intend to expand this work by releasing the complete dataset with fine-grained annotations across 14 distinct toxicity categories in the near future.

\section*{Limitations}
Due to the time and cost-intensive nature of the manual inspection, our study was constrained to six languages (and 30k test instances), as seeking qualified native-speaker annotators for a broader language set presented significant logistical challenges (Looking for Punjabi, Tamil, and Telugu speakers was extremely challenging). We implemented language code as a critical parameter in our methodology to address a fundamental limitation in toxicity detection across Indian languages: lexical items that are neutral in one linguistic context may carry offensive connotations in another. For example, particular North Indian family surnames may be perceived as toxic content by South Indian users and vice versa. Consequently, we employed language-specific parameterization to facilitate targeted model selection for individual languages rather than implementing a single unified classification model that could not account for these cross-linguistic semantic variations.

\section*{Ethics}
Our toxicity detection research prioritizes ethical considerations throughout dataset creation and model development. Annotators were informed about potentially harmful content and provided with mental health resources. We anonymized all personally identifiable information and developed culturally-sensitive annotation guidelines with native speakers to respect linguistic nuances. While our system aims to detect online harm, we acknowledge the risk of false positives potentially silencing legitimate speech, and emphasize that automated tools should supplement rather than replace human moderation in sensitive contexts.

\section*{Acknowledgments}
This work is supported by the Prime Minister Research Fellowship (PMRF-1702154) to Himanshu Beniwal. 

\bibliography{acl_latex}

\appendix

\newpage
\section{Appendix}
\label{sec:appendix}

\subsection{Dataset}
\label{sec:appendix-dataset}
Our dataset construction process involved content extraction from online sources. We employed web scraping techniques and implemented classification using a manually curated lexicon of toxic terms. Text instances containing items from this predefined vocabulary received ``toxic'' labels, while those without such terms were classified as ``non-toxic''. To simplify the annotation framework, we converted the initial multi-class labeling scheme to binary: content originally marked non-toxic retained this designation, while all gradations of toxicity (``moderately toxic'', ``highly toxic'') were consolidated into a single ``toxic'' category.

We present the language-wise source and license list as:
\paragraph{Hindi} Hindi hate speech data\footnote{\url{https://www.kaggle.com/datasets/yash3056/hindi-hate-speech-data}}, MACD\footnote{\url{https://github.com/ShareChatAI/MACD}}, textdetox/multilingual\_toxicity\_dataset\footnote{\url{https://huggingface.co/datasets/textdetox/multilingual_toxicity_dataset/viewer/default/hi}}, QCRI/LlamaLens-Hindi\footnote{\url{https://huggingface.co/datasets/QCRI/LlamaLens-Hindi/viewer/Offensive_Speech_Detection}}(Apache 2.0, CC BY-NC-SA 4.0, Creative Commons Attribution 4.0, Open Rail++-M, and MIT.).
\paragraph{Telugu} HOLD-Telugu\footnote{\url{https://github.com/Mussabat/HateSpeech-EACL-2024/}}, Detecting Insults in Social Commentary\footnote{\url{https://www.kaggle.com/c/detecting-insults-in-social-commentary/data?select=test.csv}}, and mounikaiiith/Telugu-Hatespeech\footnote{\url{https://www.kaggle.com/c/detecting-insults-in-social-commentary/}} (CC‑BY‑4.0 and CC BY-ND 3.0).
\paragraph{Marathi} MahaHate\footnote{\url{https://github.com/l3cube-pune/MarathiNLP/tree/main/L3Cube-MahaHate}}, MOLD\footnote{\url{https://github.com/TharinduDR/MOLD/tree/master/MOLD_1.0}} (CC‑BY‑4.0).
\paragraph{Urdu} Roman-Urdu-Toxic-Corpus\footnote{\url{https://huggingface.co/datasets/hafiz-hassaan-saeed/Roman-Urdu-Toxic-Corpu}}, Parallel-Urdu-Roman-Urdu-Corpus\footnote{\url{https://huggingface.co/datasets/hafiz-hassaan-saeed/Parallel-Urdu-Roman-Urdu-Corpus}}, and roman\_urdu\_hate\_speech\footnote{\url{https://github.com/haroonshakeel/roman_urdu_hate_speech}} (CC‑BY‑4.0 and MIT).
\paragraph{Punjabi} News18 Punjab, BBC News Punjabi, PTC News (CC‑BY‑4.0).
\paragraph{Tamil} CulturaX\footnote{\url{https://huggingface.co/datasets/uonlp/CulturaX/tree/main/ta}}, Thanthi\footnote{\url{https://www.dailythanthi.com/news/tamilnadu}}, and Siruvar Malar\footnote{\url{https://www.tamilsiruvarkathaigal.com/}} (MC4 License, OSCAR license, and CC BY-SA 4.0).
More details about the sources, labeling strategy, and licenses are available on our project page\footnote{\url{https://github.com/himanshubeniwal/IndicToxic}}. 

We also computed the test-set's percentage agreement and inter-annotator agreement (IAA) as shown in Table~\ref{tab:cohen-scores}. We compute Cohen's Kappa using the inbuilt function and percentage agreement by the sum of exact matches divided by a total number of instances. 
\begin{table}[t]
\centering
\begin{tabular}{ccc} 
\textbf{Language} & \textbf{Agreement} & \textbf{Cohen's Kappa} \\ \hline
\textit{Hindi} & \multicolumn{1}{c}{99.60} & \multicolumn{1}{c}{99.2} \\
\textit{Telugu} & \multicolumn{1}{c}{99.08} & \multicolumn{1}{c}{98.16} \\
\textit{Marathi} & 98 & 96.01 \\
\textit{Urdu} & \multicolumn{1}{c}{98.10} & \multicolumn{1}{c}{92.61} \\
\textit{Punjabi} & \multicolumn{1}{c}{89.02} & \multicolumn{1}{c}{77.95} \\
\textit{Tamil} & 88.97 & 83.10 \\ \hline
\end{tabular}%

\caption{Percentage agreement and Inter Annotator Agreement scores for the six languages. We report both the scores in percentages. \textbf{\textit{Takeaway}}: \textit{There is a high agreement that the content is actually toxic.}}
\label{tab:cohen-scores}
\end{table}

\begin{figure}[t]
    \centering
    \includegraphics[width=\linewidth]{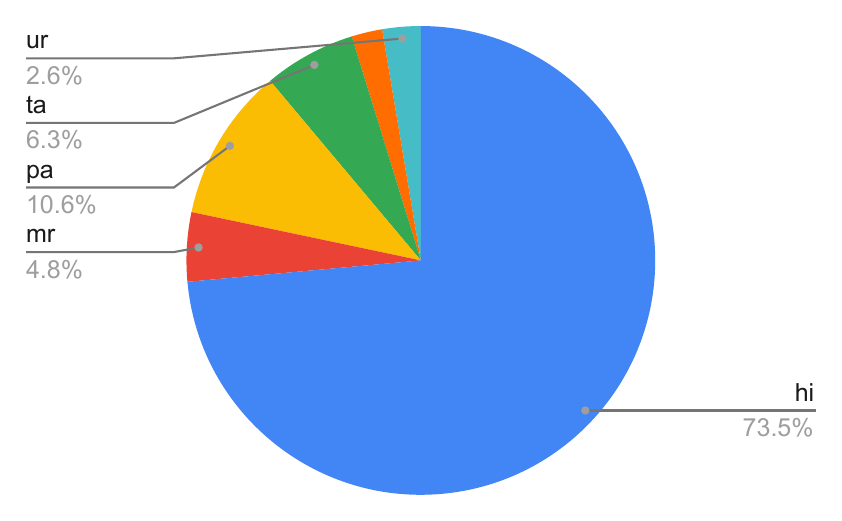}
    \caption{The proportion of languages inferred in the UI interface. \textbf{\textit{Takeaway}}: \textit{Hindi has been inferred the maximum}.}
    \label{fig:languageinfer}
\end{figure}

\begin{figure}[t]
    \centering
    \includegraphics[width=\linewidth]{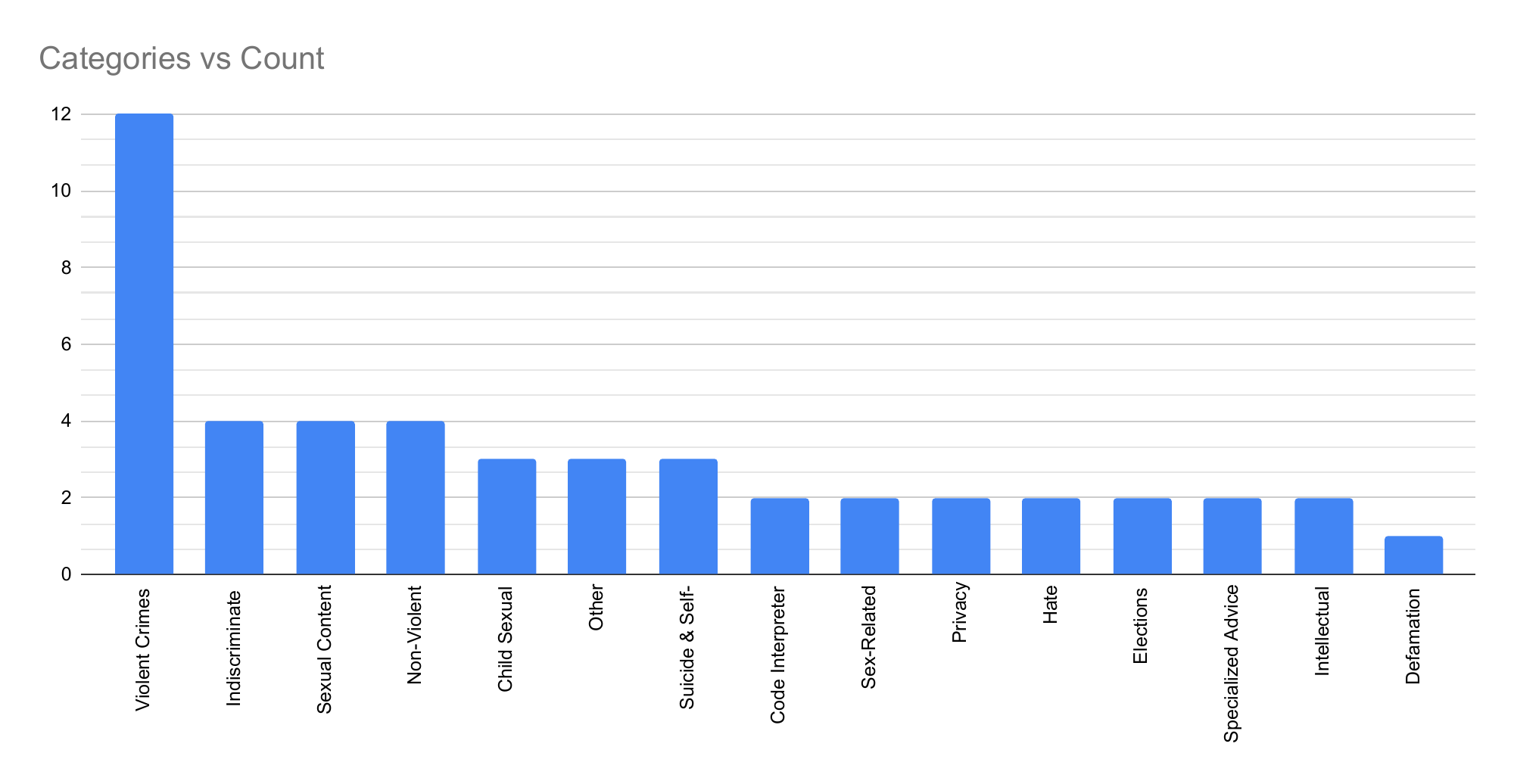}
    \caption{Categories reported by users during the feedback system. \textbf{\textit{Takeaway}}: \textit{Majority of the violent crimes were reported by the users}.}
    \label{fig:feedbackcategroeis}
\end{figure}

\section{Inception Since Launch}
\label{sec:inception}
Since the framework's launch in last week of March 2025, the UI has recorded over 200 requests. Figure~\ref{fig:languageinfer} presents the language distribution, showing the highest inference frequency for Hindi (``\textit{hi}''), while low-resource languages such as Tamil (``\textit{ta}'') and Urdu (``\textit{ur}'') exhibited minimal usage. Toxicity analysis revealed that 33\% of responses were flagged as toxic, with 66.7\% classified as non-toxic. Among predicted labels, 58.1\% of users reported relevant generation quality. Notably, toxicity definitions may vary across users. Figure~\ref{fig:feedbackcategroeis} displays user-reported feedback categories.

\end{document}